\documentclass[journal,twoside]{IEEEtran}
\IEEEoverridecommandlockouts
\usepackage{cite}
\usepackage{amsmath,amssymb,amsfonts}
\usepackage{algorithmic}
\usepackage{graphicx}
\usepackage{textcomp}
\usepackage{xcolor}
\usepackage{hyperref}
\def\BibTeX{{\rm B\kern-.05em{\sc i\kern-.025em b}\kern-.08em
    T\kern-.1667em\lower.7ex\hbox{E}\kern-.125emX}}
\begin{document}

\markboth{IEEE ROBOTICS AND AUTOMATION LETTERS. PREPRINT VERSION. SUBMITTED JULY, 2026}
{Moncada-Ramirez \MakeLowercase{\textit{et al.}}: Human-Robot Interaction in GenAI Architectures via the Agent-Client Protocol}

%%%%%%%%%%%%%%% CUSTOM COMMANDS %%%%%%%%%%%%%%%
\providecommand{\EQ}[1]{Eq.~(\ref{#1})}
\providecommand{\FIG}[1]{Fig.~\ref{#1}}
\providecommand{\SEC}[1]{Section~\ref{#1}}
\providecommand{\SECS}[2]{Sections~\ref{#1}--\ref{#2}}
\providecommand{\APPX}[1]{Appendix~\ref{#1}}
\providecommand{\TABLE}[1]{Table~\ref{#1}}
\providecommand{\MATRIX}[1]{\mathbf{#1}}
\newcommand{\etal}{et~al.~}
\newcommand{\ie}{i.e.,~}
\newcommand{\eg}{e.g.,~}
\newcommand{\raul}[1]{\colorbox{cyan}{\color{white}   \textsf{\textbf{Raul}}} \textcolor{cyan}{#1}}
\newcommand{\jesus}[1]{\colorbox{orange}{\color{white}   \textsf{\textbf{Jesus}}} \textcolor{orange}{#1}}
\newcommand{\javier}[1]{\colorbox{purple}{\color{white}   \textsf{\textbf{Javier}}} \textcolor{purple}{#1}}
\newcommand{\cipri}[1]{\colorbox{purple}{\color{white}   \textsf{\textbf{Cipri}}} \textcolor{purple}{#1}}
\newcommand\red[1]{\textcolor{red}{#1}}
\newcommand{\todo}[1]{\colorbox{red}{\color{white}   \textsf{\textbf{TODO}}} \textcolor{red}{#1}}
\newcommand{\review}[1]{\colorbox{orange}{\color{white}   \textsf{\textbf{REVIEW}}} \textcolor{orange}{#1}}
\newcommand{\discuss}[1]{\colorbox{magenta}{\color{white}   \textsf{\textbf{DISCUSS}}} \textcolor{magenta}{#1}}
\newcommand{\summarizable}[1]{\colorbox{teal}{\color{white}   \textsf{\textbf{SUMMARIZABLE}}} \textcolor{teal}{#1}}
\newcommand{\add}[1]{\colorbox{green}{\color{white}   \textsf{\textbf{ADD}}} \textcolor{green}{#1}}
\newcommand{\remove}[1]{\colorbox{red}{\color{white}   \textsf{\textbf{REMOVE}}} \textcolor{red}{#1}}

\title{Human-Robot Interaction in GenAI Architectures via the Agent-Client Protocol}

\author{Jesus Moncada-Ramirez\textsuperscript{1}, Jose-Raul Ruiz-Sarmiento\textsuperscript{1}, and Javier Gonzalez-Jimenez\textsuperscript{1}%
\thanks{This work has been submitted to the IEEE for possible publication. Copyright may be transferred without notice, after which this version may no longer be accessible.}%
\thanks{\textsuperscript{1}The authors are with the Machine Perception and Intelligent Robotics Group (MAPIR-UMA), Malaga Institute for Mechatronics Engineering and Cyber-Physical Systems (IMECH.UMA), University of Malaga, 29071 Málaga, Spain (e-mail: \texttt{jemonra@uma.es}; \texttt{jotaraul@uma.es}; \texttt{javiergonzalez@uma.es}).}%
}

\maketitle

\begin{abstract}
% Intro
Recent advances in Generative Artificial Intelligence (GenAI), particularly Large Language Models (LLMs), are driving robotic architectures toward agent-based high-level orchestration, in which natural-language instructions can be translated into context-aware action sequences.
% Problem
While the integration of these agents and robotic capabilities is increasingly converging toward standardization through the Model Context Protocol (MCP), the upper Human-Robot Interaction (HRI) layer remains fragmented by proprietary, ad hoc interfaces that hinder real-time human-in-the-loop collaboration. 

% Solution -> include ACP
To address this fragmentation, this paper proposes the adoption of the Agent-Client Protocol (ACP)---a communication standard originally introduced for coding agents in software engineering---as a unified communication contract for the HRI layer in agent-based robotic systems. 
% Three-layer architecture
By combining ACP at the interface-agent link and MCP at the agent-execution link, we formulate a fully decoupled three-layer architecture that separates human interaction, deliberative orchestration, and physical execution.
% Advantages 1
This topology removes rigid architectural dependencies, enabling heterogeneous user interfaces to connect to the same robotic system and allowing the underlying robotic platform to be replaced without requiring client-specific integration changes. 
% Advantages 2
Moreover, it provides native support for collaborative HRI capabilities such as real-time observability, explicit human authorization, and immediate task interruption.
% Experiments
We experimentally evaluate the proposed architecture on a physical mobile robot, demonstrating interoperability across three heterogeneous user interfaces and validating real-time human-in-the-loop workflows with negligible latency overhead.
\end{abstract}


\section{Introduction}
\label{sec:introduction}
% From traditional robotics to GenAI robotics
While robotics has traditionally relied on tightly coupled software architectures and deterministic logic systems for high-level decision-making and planning~\cite{siciliano2008springer}, recent advances in Generative Artificial Intelligence (GenAI) are reshaping this paradigm~\cite{firoozi2025foundation}.
In particular, the adoption of Large Language Models (LLMs)~\cite{ahn2022can} and Vision-Language-Action (VLA) models~\cite{zitkovich2023rt} is driving robotic systems toward architectures based on autonomous AI agents, where high-level natural-language instructions can be translated into context-aware action sequences.
This evolution alleviates some of the rigidity of traditional programming abstractions---such as finite state machines and behavior trees---by enabling robots to interpret intuitive user instructions (\eg \textit{``Carry the medical box to room 3''}) and reason about task execution at a semantic level.
As a result, these models extend robotic systems with capabilities such as common-sense reasoning, contextual generalization, and real-time dynamic replanning.

\begin{figure}[t]
    \centering
    \includegraphics[width=\columnwidth]{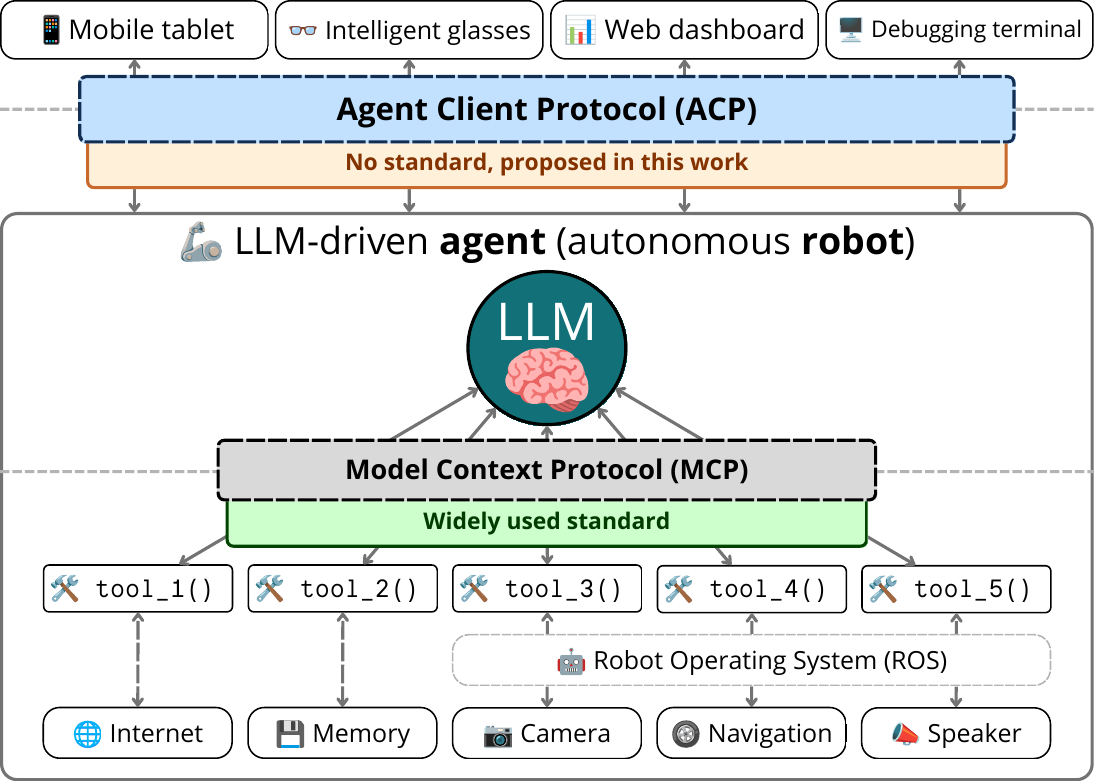}
    \caption{Architecture overview of an autonomous robot powered by LLMs.
    In the lower execution layer, the agent acts as a client to an MCP server to consume tools that grant access to hardware (such as the camera via ROS) and other external services (Internet). 
    In the upper layer, the ACP unifies bidirectional communication between the robot (agent) and various user interfaces (clients).}
    \label{fig:scenario}
    \vspace{-0.4cm}
\end{figure}

% GenAI robotics working
To implement this form of high-level deliberation, a common architectural pattern is to employ an LLM as a central orchestrator, equipped with a set of \emph{tools} that provide access to the robot's primitive capabilities (\eg perception, navigation, or manipulation)~\cite{ahn2024autort, song2023llm}.
This integration of LLMs with external functionalities defines the operational framework of an autonomous agent, typically structured under the \textit{Reason+Act} (ReAct) paradigm~\cite{yao2022react}.
Under this model, the agent interleaves internal reasoning with the execution of \emph{function calls}, forming an iterative cycle of thought, action, and observation that enables closed-loop deliberative control.
For example, given the instruction \textit{``Pick up the battery''}, the agent may first generate an intermediate reasoning step (\textit{``I must locate the battery visually''}), invoke a perception tool (\texttt{detect\_objects()}), and use the resulting observation (\textit{``The battery is at coordinates X, Y''}) to guide its next action (\texttt{navigate\_to(X, Y)}).

To formalize the integration between LLM-driven agents and their available tools, the \emph{Model Context Protocol} (MCP)~\cite{anthropic2024mcp} has emerged as an open standard for tool discovery and invocation.
Architecturally, MCP follows a client-server model in which the server encapsulates the functionalities of an underlying system and exposes them to the LLM through a standardized communication contract.
In the robotics domain, an MCP server can encapsulate the robot's execution middleware---such as Robotic Operating System (ROS) topics, services, or actions---and expose it to the agent as a structured catalog of tools.

% Problems in the interaction layer
Although the interaction between LLM-driven agents and robotic execution systems is progressively converging toward standardization through MCP, the same does not hold for the upper interaction layer.
The bidirectional communication between human operators and robotic agents remains fragmented across proprietary and tightly coupled implementations~\cite{vemprala2023chatgpt}.
This lack of a common communication contract imposes substantial integration overhead, requiring user interfaces to be redesigned or refactored whenever the robotic system changes, or conversely, forcing modifications in the robot's high-level communication logic when deploying new interfaces.
Beyond engineering costs, this fragmentation also limits Human-Robot Interaction (HRI).
On one hand, it obscures the agent's intermediate reasoning and decision-making process, making it difficult for users to inspect the system's internal state.
On the other hand, it complicates critical collaborative operations such as requesting explicit user authorization for sensitive actions or interrupting an ongoing task to provide corrective instructions in real time.
Ultimately, this lack of standardization hinders the realization of effective human-in-the-loop robotic workflows.

% ACP: problem solved in software engineering
A similar fragmentation problem has already been addressed in the software engineering domain.
Early AI-driven programming assistants---now commonly referred to as \emph{coding agents} (\eg Claude Code, Codex)---relied on proprietary integrations tied to specific Integrated Development Environments (IDEs) (\eg VS Code, IntelliJ), resulting in duplicated integration effort across platforms.
To address this limitation, the community recognized the need to decouple the user interface (the client) from the AI logic (the agent), establishing a common communication layer for transmitting user intent, managing permissions, and visualizing the agent's progress in real time.
This led to the creation of the \emph{Agent-Client Protocol} (ACP)~\cite{acp2026standard}, a bidirectional standard that allows any application (client) to interact with AI agents (servers) through a unified and platform-agnostic communication contract.

Motivated by this standardization strategy, this work proposes and experimentally evaluates the adoption of ACP as the communication protocol for the upper interaction layer in GenAI-orchestrated robotic systems.
By combining ACP at the interface--agent link and MCP at the agent--execution link, we formulate a fully decoupled three-layer architecture.
This topology removes rigid architectural dependencies and incorporates human interaction as a native component of the system, enabling real-time observability, ambiguity resolution, and explicit permission management.

% Scenario
To illustrate the architectural implications of this proposal, consider a mobile manipulator powered by an LLM-based agent tasked with inspecting and classifying components on a factory floor (see \FIG{fig:scenario}).
Under a non-unified architecture, enabling human supervision requires implementing custom communication pathways for each user interface.
For example, if operators need to monitor the robot through a web dashboard, a mobile tablet, and a developer terminal, each frontend must be individually integrated with the robot's backend communication logic, often requiring additional \textit{ad hoc} mechanisms for state transmission, intervention, or assistance requests.
By contrast, encapsulating the robotic agent behind the ACP exposes a single standardized communication contract.
This allows heterogeneous ACP-compatible clients to connect to the same robotic system without requiring modifications to the robot-side orchestration logic.
As a result, ACP not only reduces integration effort but also provides a common substrate for advanced HRI capabilities across multiple interfaces.

% Evaluation
To evaluate the architectural feasibility of the proposal, we deployed the architecture on a physical mobile robot called Sancho.
We validated the connection of three heterogeneous user interfaces (a developer CLI, a tablet application, and a generic web client) without requiring modifications to the robot's orchestration layer.
From the experiments conducted, we report a representative assistance session and analyze its execution trace, illustrating how the architecture natively supports real-time human-in-the-loop interaction.
Furthermore, latency analysis of representative HRI workflows shows that the communication overhead introduced by ACP ($1.0$~ms) and MCP ($170.6$~ms) is negligible, amounting to only $171.6$~ms (less than $0.5$\% of the average $36.9$-second interaction loop, which is dominated by LLM inference and physical tool execution such as navigation).
To promote reproducibility, the source code implementing the proposed architecture on Sancho\footnote{Sancho source code: \url{https://github.com/MAPIRlab/sancho_robot/}} and the developed client interfaces\footnote{ACP clients developed: \url{https://github.com/MAPIRlab/sancho-acp-clients}} have been released as open-source.

\section{Related Work}

The transition toward agent-based robotic architectures has gained significant attention in recent years, largely driven by the emergence of LLMs and multimodal foundation models.
This section reviews the current landscape of AI-driven robotics, focusing on orchestration paradigms, hardware abstraction, and the remaining challenges in HRI standardization.

\subsection{LLM-Driven Robotic Orchestration}

% Emergence
Early integrations of LLMs into robotics exploited their zero-shot reasoning capabilities to translate abstract natural-language instructions into structured task decompositions~\cite{huang2022language}.
To improve the physical feasibility of these plans, frameworks such as SayCan~\cite{ahn2022can} introduced grounding mechanisms to align the model's output with the robot's available affordances.
Building upon these foundations, architectures such as LLM-Planner~\cite{song2023llm} extended the paradigm through hierarchical planning, using the LLM to generate high-level sub-goals that were subsequently executed by specialized low-level controllers.
% VLAs
Beyond purely text-driven architectures, the field has progressively evolved toward Vision-Language-Action (VLA) models, such as RT-1~\cite{brohan2022rt} and RT-2~\cite{zitkovich2023rt}, which provide a more direct mapping from multimodal observations and user intent to low-level control policies.

% Evolution 
While static planning and purely reactive strategies remain effective for short-horizon tasks, more complex and open-ended scenarios require foundation models to operate as high-level orchestrators within closed-loop control systems.
Recent architectures exemplify this transition by placing the foundation model at the center of iterative perception-action loops~\cite{moncada2025agentic}.
For instance, AutoRT~\cite{ahn2024autort} adopts this paradigm to coordinate fleets of robots in unstructured environments, while SayPlan~\cite{rana2023sayplan} grounds the model in 3D scene graphs to support long-horizon task execution.
Inspired by iterative reasoning frameworks such as ReAct~\cite{yao2022react} and Inner Monologue~\cite{huang2022inner}, these systems allow the LLM to continuously observe the environment, perform actions, and replan in response to execution failures.

% Our work
Despite these advances in autonomy, their implementations typically remain tightly coupled to specific hardware stacks or software ecosystems, lacking a unified abstraction layer for interoperability across heterogeneous robotic platforms.
To address this limitation, the architecture proposed in this work isolates the agent orchestrator into a fully decoupled central layer.
By relying exclusively on standardized communication contracts, the proposed approach ensures that the agent's reasoning remains independent of both the underlying execution platform and the external user interfaces.

\subsection{Hardware Abstractions}

To perform tasks beyond text generation, LLM-driven agents require structured interfaces to discover and interact with external tools and APIs.
In the robotics domain, early integration approaches, such as Code as Policies~\cite{liang2023code}, leveraged the code-generation capabilities of LLMs to synthesize executable Python programs that directly invoked hardware APIs.
Other works, such as ProgPrompt~\cite{singh2022progprompt}, structured this interaction by exposing the model to a predefined set of high-level callable functions.
Despite these advances, the discovery, invocation, and documentation of these capabilities have traditionally relied on \textit{ad hoc} parsers and custom integrations.

To address this fragmentation, recent efforts have moved toward protocol-level standardization through the MCP~\cite{anthropic2024mcp}.
By adopting a client-server architecture, MCP defines a unified JSON-RPC communication contract for tool discovery and execution, abstracting the complexities of the underlying robotic middleware~\cite{guan2025roboneuron, ros2mcp_community}.
% Our work
While our work adopts MCP for the lower execution layer, we note that standardizing the hardware link only solves half of the architectural problem, leaving the human-interaction layer unaddressed.

\subsection{Human-Robot Interaction in GenAI Architectures}

Despite recent progress in agent-based robotic systems, the bidirectional communication between human operators and LLM-driven robots remains largely implemented through task-specific and tightly coupled interfaces~\cite{ren2023robots, cui2023no}.
Current HRI solutions in GenAI-enabled robotics typically rely on proprietary communication channels and application-specific frontends~\cite{vemprala2023chatgpt}.
While traditional robotic visualization tools (\eg RViz, Foxglove) provide effective support for monitoring deterministic system telemetry (\eg point clouds, coordinate transforms, or navigation states), they lack the semantic abstractions required to support natural-language interaction and expose the internal reasoning processes of LLM-based agents.
As a result, functionalities that are increasingly important for collaborative embodied AI---such as real-time intent visualization, explicit human authorization for sensitive actions, and context-preserving session management---must often be implemented independently for each robotic system~\cite{cui2023no}.
This fragmentation limits interoperability and increases integration complexity across heterogeneous interfaces.

As discussed earlier, the software engineering domain addressed a similar fragmentation problem between coding agents and IDEs through the ACP~\cite{acp2026standard}.
However, to the best of our knowledge, no prior work has adapted and evaluated this decoupling strategy in the robotics domain.
The architecture proposed in this work addresses this gap by introducing ACP as a common communication protocol for the upper interaction layer, enabling heterogeneous user interfaces to interact with robotic agents in a decoupled, unified manner.

\begin{figure*}[!ht]
\centering
    \includegraphics[width=1.0\textwidth]{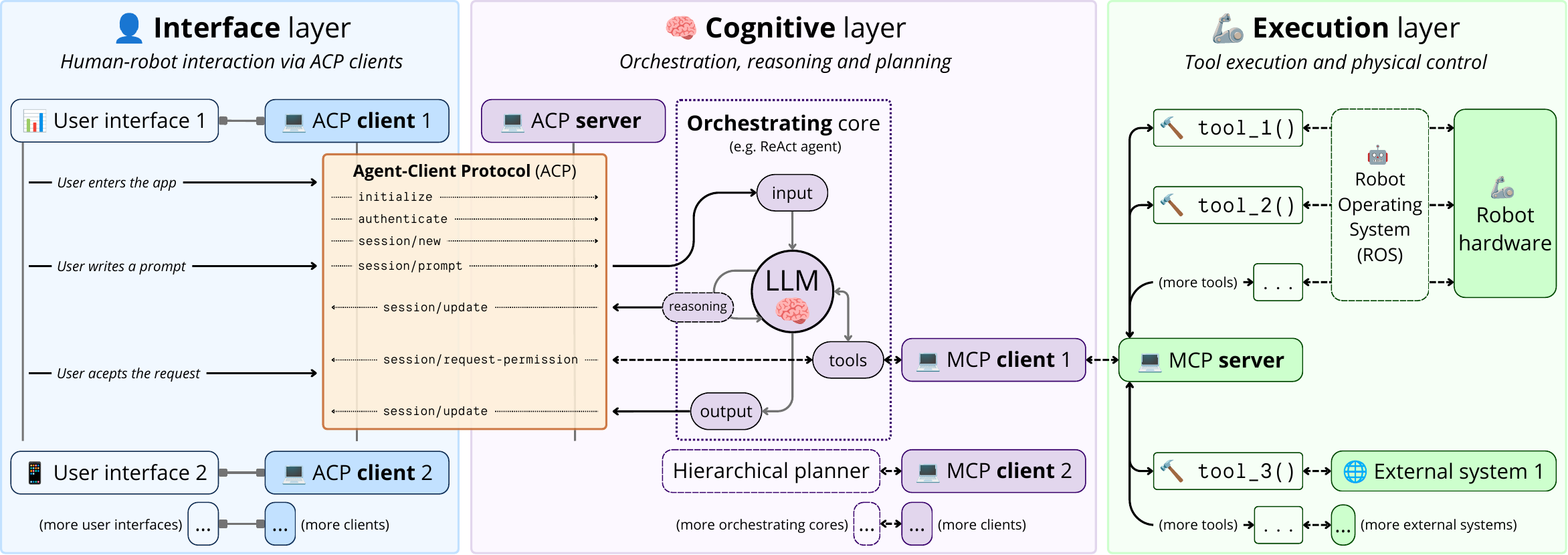}\\
    \caption{Overview of the proposed robotic architecture.
    The architecture is structured into three communicating layers: 
    an interface layer composed of multiple ACP clients for HRI,
    a cognitive layer that operates as an ACP server and an MCP client to perform orchestration and planning,
    and an execution layer that exposes robot and external capabilities through an MCP server while ensuring deterministic control and safety.}
    \label{fig:pipeline}
    
\end{figure*}

\section{Description of the Proposed Architecture}

% Overview
The proposed robotic architecture, illustrated in \FIG{fig:pipeline}, is a fully decoupled three-layer system connected through asynchronous JSON-RPC communication contracts: ACP at the upper interface--agent link and MCP at the lower agent--execution link.
Functionally, the interface layer governs HRI (see \SEC{subsec:interface_layer}), the cognitive layer concentrates high-level orchestration and planning (see \SEC{subsec:cognitive_layer}), and the execution layer encapsulates physical control, deterministic safety constraints, and hardware-level operations (see \SEC{subsec:execution_layer}).

\subsection{Interface Layer}
\label{subsec:interface_layer}

% Overview and naming
The interface layer represents the user-facing entry point (frontend) of the architecture and can be instantiated through heterogeneous ACP \emph{clients}, \ie user interfaces such as a console, a web dashboard, or a mobile application, without requiring modifications to the robotic \emph{agent} hosted in the cognitive layer, which operates as the ACP server.
Under the ACP, and adapted here to the robotics domain, the client is an interaction layer not responsible for planning or physical control of the robot.
All deliberation resides in the agent (server), while the client is limited to managing the dialogue session, capturing user intents, and visualizing the agent's feedback.

% Sessions?
In this context, a \emph{session} is defined as a continuous, stateful communication lifecycle between the client interface and the robotic agent.
Rather than a single prompt-response exchange, a session spans multiple dialogue turns and iterative ReAct loops, encompassing user instructions, tool executions, environment observations, and dynamic replanning.
It acts as an interaction context, preserving the agent's short-term memory and operational state until the connection is explicitly terminated by the client.

The following list summarizes the client-side operations defined by the ACP and describes their practical role within the proposed robotic architecture.

\paragraph{Connection Initialization}
% Initialization to create communication channel
At the start of the interaction, the client and the agent establish a communication channel to ensure protocol compatibility and validate the client's identity (through the \texttt{initialize} operation in the ACP specification).
During this step, both parties negotiate their supported capabilities, including client-side resources (\eg file system access, terminal execution) and agent-side functionalities (\eg multimodal prompts or MCP connectivity), which define the operational scope of subsequent sessions.
This initialization model also enables concurrent interaction, allowing multiple client interfaces to connect simultaneously to the same robotic system.

\paragraph{Authentication}
To guarantee access control, the agent may require credential validation (\texttt{authenticate}), ensuring that only clients with the appropriate authorization can establish sessions that affect the robot's operation.

\paragraph{Session Management}
The interaction is structured through a complete lifecycle that allows creating (\texttt{session\_new}), loading (\texttt{session\_load}), listing (\texttt{session\_list}), forking (\texttt{session\_fork}), resuming (\texttt{session\_resume}), and closing sessions (\texttt{session\_close}).
This management is vital at an operational level, as it ensures that the robot agent does not lose context when switching sessions, allows for immediate control recovery after a network failure, and ensures a traceable log of all decisions made, among others.

\paragraph{Configuration Selection}
The behavior of the robotic agent during a session can be dynamically modified by selecting the appropriate operating mode (\texttt{session/set\_mode}), which allows adjusting the agent's available tools, system prompts, and permission requirements on the fly.
Translated to a robotic context, this feature allows seamless transitions between passive (\eg stationary sensor monitoring, reasoning, trajectory generation without physical actuation) and active profiles (\eg movement-only, fully autonomous operation, human-in-the-loop supervised execution).
Additionally, the client can adjust the LLM governing the agent (\texttt{session/set\_model}), choosing the most suitable one depending on cost, response speed, and other requirements of the intended task.
Beyond these, the agent can publish additional settings to be configured by the client (\texttt{session/set\_config\_option}).

\paragraph{Instruction Delivery}
The core of the interaction materializes when the operator transmits directives in natural language, optionally supported by multimodal context (\texttt{session/prompt}).
This input triggers the agent's cognitive cycle, which processes the user intent, plans the necessary physical actions, and asynchronously reports its progress within the session flow.

\paragraph{Cancellation}
To guarantee dynamic and safe human intervention, the protocol incorporates an immediate interruption mechanism (\texttt{session/cancel}).
This operation allows the operator to halt the cognitive loop and physical execution in real-time in the face of operational risks, context degradation, or the need to replan the mission.

Reverse operations, initiated by the server toward the client, are detailed in \SEC{subsec:cognitive_layer}, as they originate in the agent's orchestration logic.

\subsection{Cognitive Layer}
\label{subsec:cognitive_layer}

The cognitive layer forms the deliberative core of the architecture, where an LLM transforms user intent into high-level action plans under the coordination of an orchestrator core (see \SEC{subsubsec:orchestrating_core}).
Operationally, this layer maintains two simultaneous communication links: it operates as an ACP server toward the interface layer (see \SEC{subsubsec:acp_server}) and as an MCP client toward the execution layer for tool discovery and invocation (see \SEC{subsubsec:mcp_client}).

\subsubsection{Orchestrator Core} \label{subsubsec:orchestrating_core}
The orchestrator core concentrates the deliberative logic of the system by transforming user intent into executable action sequences.
Although it typically follows the ReAct~\cite{yao2022react} paradigm, its design is compatible with other orchestration paradigms, including hierarchical planning and multi-agent coordination.
At each iteration, the orchestrator evaluates the current context, invokes tools through the MCP client, and integrates the resulting observations to refine its action plan.
In parallel, it uses the ACP to stream its intermediate reasoning and execution progress to the user interface, enabling clarification requests or human validation when ambiguities or execution failures arise.

Overall, the orchestrator maintains an iterative loop of reasoning, execution, and evaluation.
This enables the system to adapt safely to unforeseen events while preserving a strict separation between high-level deliberation and low-level physical control.

\subsubsection{ACP Server} \label{subsubsec:acp_server}

Within the proposed architecture, the orchestrator core leverages the ACP server to issue requests and notifications to the ACP client (recall \SEC{subsec:interface_layer}).
The following list summarizes the server-initiated communication primitives defined by the ACP, highlighting their role in enabling real-time observability, human-in-the-loop validation, and feedback during robotic execution.

\paragraph{Status Updates to the Client}
Real-time supervision is guaranteed through a continuous stream of events reporting on agent messages, thoughts, and tool usage (\texttt{session/update}).
In this way, the operator maintains visibility into the agent's internal state and can interpret how the task evolves at each stage.

\paragraph{Human Authorization Requests}
Human intervention is explicitly integrated when execution requires prior validation for a sensitive action (\texttt{session/request\_permission}).
This control point formalizes the human-in-the-loop scheme and reduces operational risk in decisions impacting safety, the environment, or the mission.

\paragraph{Local File Read and Write}
Although not a strictly central capability in many robotic deployments, reading and writing files (\texttt{fs/read\_text\_file}, \texttt{fs/write\_text\_file}) can be useful for the agent to transfer files to the client, such as execution traces or diagnostic logs

\paragraph{Terminal Process Management}
The terminal process management on the client provided by the ACP (\texttt{terminal/create}, \texttt{terminal/output}, \texttt{terminal/wait\_for\_exit}, \texttt{terminal/release}, \texttt{terminal/kill}) generally makes little sense in a robotic deployment; therefore, these operations should be restricted.

\subsubsection{MCP Client} \label{subsubsec:mcp_client}

The MCP client bridges the cognitive layer with the robot's physical capabilities by consuming the formal catalog of structured tools (\eg navigate to a waypoint, open the gripper, check the battery level) exposed by the execution layer (see \SEC{subsec:execution_layer}).
The LLM itself does not interact with the hardware; rather, it outputs tool-call intents which the MCP client forwards to the MCP server for execution, returning the results as contextual feedback to the model, always under the control of the orchestrator core (recall \SEC{subsubsec:orchestrating_core}).
This separation decouples cognitive logic from physical implementation, ensuring portability across different robotic platforms.

\subsection{Execution Layer}
\label{subsec:execution_layer}

The execution layer establishes the connection between high-level orchestration and the physical operation of the robot through an MCP server.
This component encapsulates the operational capabilities of the system---including native middleware functionalities (such as ROS topics, services, and actions) as well as external resources (such as web services or cloud queries)---and exposes them to the cognitive layer as a unified catalog of standardized tools.

From an engineering perspective, this layer retains full responsibility for low-level control and operational safety.
The MCP server validates invocation parameters, enforces physical constraints, and rejects unsafe actions before they reach the hardware.
Consequently, non-deterministic generative inference is isolated from fast control loops, leaving stability, temporal guarantees, and emergency-stop mechanisms to specialized deterministic components.

\section{Experimental Evaluation}

This section presents the experimental evaluation of the proposed architecture through its deployment on a physical mobile robot operating in a standard office environment (see \SEC{subsec:evaluation_setup}).
The evaluation focuses on three architectural properties.
First, we assess the decoupling between the robotic agent and its user interfaces enabled by the ACP, demonstrating that three heterogeneous clients can connect to the robot without requiring modifications to its orchestration logic (see \SEC{subsec:decoupling_robotic_agent_user_interfaces}).
Second, we present a representative collaborative assistance scenario and analyze its execution trace within the architecture, illustrating how the proposed design supports real-time human-in-the-loop interaction (see \SEC{subsec:hri_workflow_validation}).
Finally, we analyze the computational overhead introduced by the architectural decoupling through ACP and MCP, showing that the proposed three-layer design preserves the robot's real-time responsiveness (see \SEC{subsec:overhead_analysis}).

\subsection{Evaluation Setup}
\label{subsec:evaluation_setup}

% Sancho!
The proposed three-layer architecture has been implemented and deployed on the mobile robot \emph{Sancho}, which is based on the ROS 2 Humble framework.
% Hardware description
Built upon an AgileX Ranger Mini V3 base, Sancho is equipped with two Hokuyo laser scanners to support robust navigation and reactive capabilities.
For environmental perception and HRI, it incorporates an Orbbec Astra RGB-D camera, as well as a head-mounted RGB camera and an integrated speaker for verbal communication and social interaction.

% Execution Layer
To materialize the execution layer, an MCP server was developed to interface with the robot's hardware and middleware.
Given that all physical capabilities of the robot are natively managed by ROS, this server acts as a semantic bridge to the underlying ROS topics, services, and actions.
The defined MCP tools enable the agent to get a pre-built topological map of the environment, navigate to specific coordinates or locations, capture images via the RGB camera, and synthesize speech through the robot's integrated speakers.

% Cognitive Layer
In the cognitive layer, the orchestrator core is implemented as a pure ReAct agent driven by the LLM \texttt{gemini-3.1-flash-lite}.
Given the user's prompt, this agent iteratively reasons and invokes the available MCP tools until the task is completed.
As a specific implementation detail, an initial LLM call is executed to generate the robot's preliminary thought.
Similar to how many conversational interfaces display their thought process while generating a longer response, this initial intention is shared with the ACP clients as the very first message, providing users with immediate feedback on what the robot is about to do.
Regarding the ACP server in this layer, since the ACP currently relies on standard input/output (\texttt{stdio}) for local process communication, the protocol was deployed over a raw TCP transport layer to enable remote network connection, requiring clients to establish a socket connection before exchanging JSON-RPC messages.

% Interface Layer
In the interface layer, three heterogeneous user interfaces were deployed to act as independent ACP clients. 
A description of them, along with the demonstration of how the architecture seamlessly accommodates their connection without requiring any custom integration adapters or modifications to the robot's logic, is discussed below.

\subsection{Decoupling of the Robotic Agent and User Interfaces}
\label{subsec:decoupling_robotic_agent_user_interfaces}

\begin{figure}[t]
    \centering
    \includegraphics[width=\columnwidth]{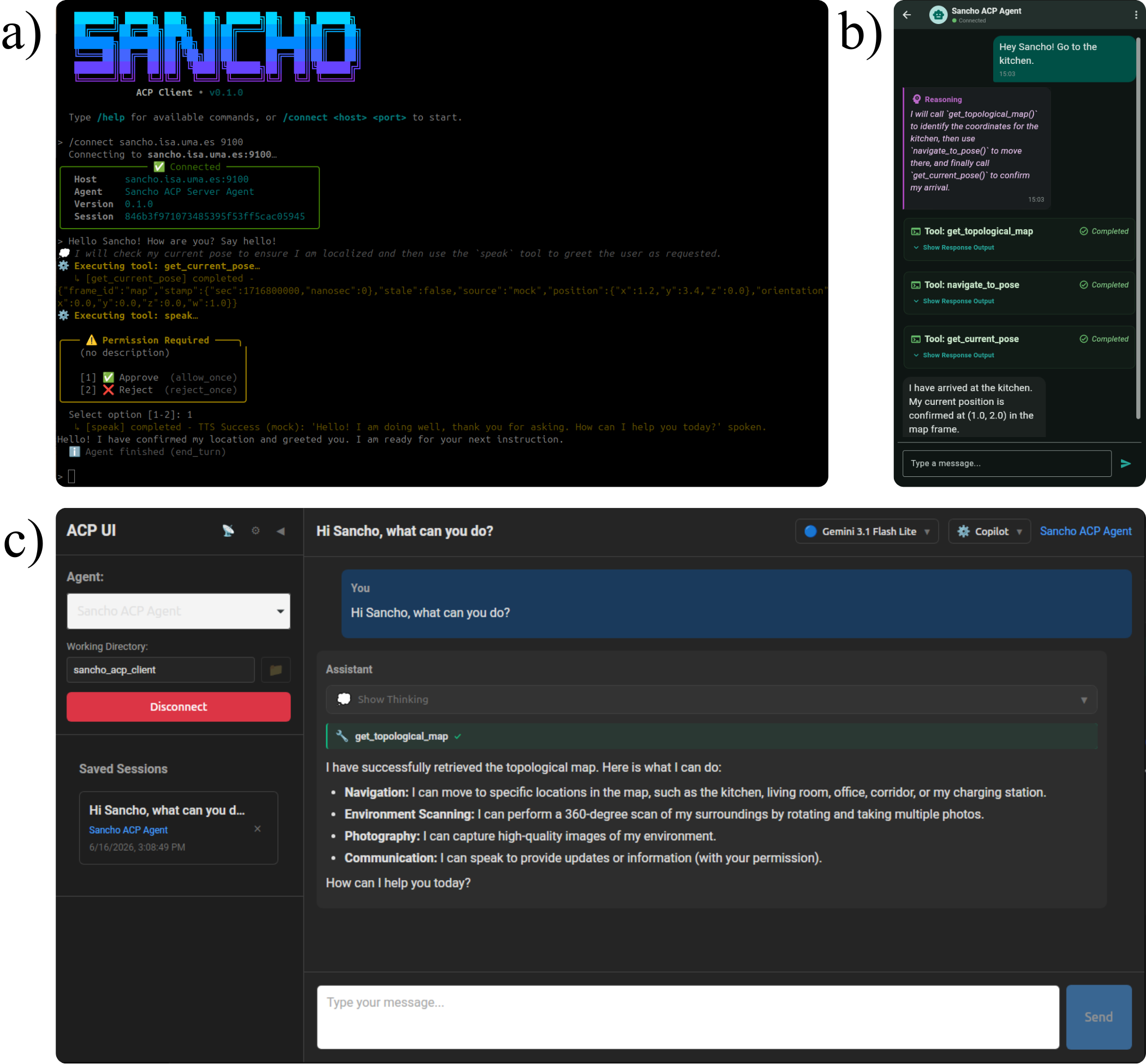}
    \caption{The three heterogeneous ACP clients connected to Sancho in the experiments: (\textbf{a}) developer CLI, (\textbf{b}) mobile app, and (\textbf{c}) \texttt{acp-ui}}
    \label{fig:interfaces_and_robot}
    \vspace{-0.4cm}
\end{figure}

% Objective
To validate that the proposed architecture successfully decouples the robot (agent, whose logic lives in the cognitive layer) from the user interfaces (clients, which live in the interface layer), 
% What we did
we connected three different ACP clients to Sancho without modifying its orchestrator (see \FIG{fig:interfaces_and_robot}).
Specifically, the evaluated interfaces include two custom-developed clients and one pre-existing third-party application:
(i)~a Developer CLI (see \FIG{fig:interfaces_and_robot}.a), our custom minimalist terminal-based client for low-level command injection;
(ii)~a Mobile app (see \FIG{fig:interfaces_and_robot}.b), our custom touchscreen-based interface deployed on a tablet for on-site operators; and
(iii)~\texttt{acp-ui} (see \FIG{fig:interfaces_and_robot}.c)\footnote{\url{https://github.com/formulahendry/acp-ui}}, a pre-existing, general-purpose open-source client.

% Outcome
Integrating these diverse clients required no custom code on the robot, thanks to the implementation of the ACP in the robot (ACP server in the cognitive layer) and user interfaces (ACP clients in the interface layer), demonstrating how this protocol enables true decoupling in the architecture.
Actually, not only does it allow multiple interfaces to connect to a single robot, but conversely, it ensures that the developed user interfaces can instantly connect to any other robotic system implementing the ACP, showcasing the complete portability and independence of the interface layer.

\subsection{HRI Workflow Validation}
\label{subsec:hri_workflow_validation}

% Objective
To evaluate the architecture's suitability for HRI, 
% What we did
several experimental trials were conducted; among these, we highlight a representative usage session of Sancho interacting with a user via the \texttt{acp-ui} client in an assistance robotic context.

\begin{figure}[t]
    \centering
    \includegraphics[width=\columnwidth]{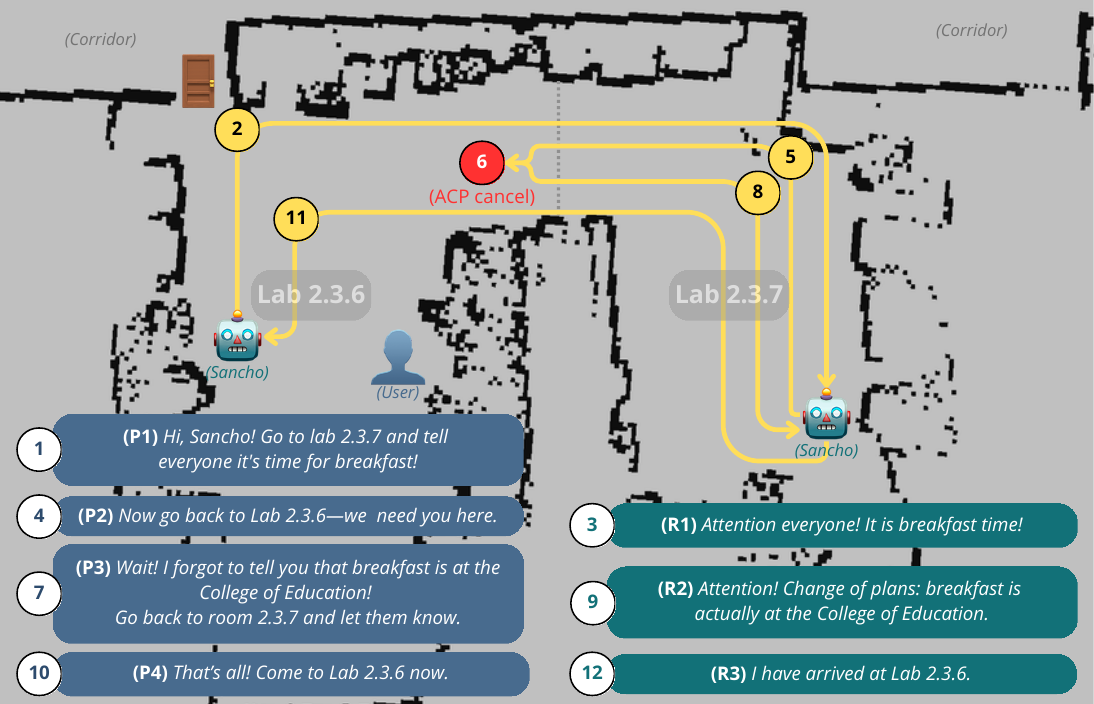}
    \caption{Map of the experimental environment showing the robot's trajectory and interaction sequence of the HRI session.
    Numbered circles indicate the chronological execution order: white circles represent user prompts (P1--P4) and robot responses (R1--R3), yellow circles represent navigation actions, and the red circle (6) denotes a user-triggered cancellation through ACP.}
    \label{fig:experimental_map}
    \vspace{-0.4cm}
\end{figure}

% Session explained
The trace of this representative session, depicted in \FIG{fig:experimental_map}, unfolded as follows.
First, the user instructed Sancho to navigate to a specific laboratory location and deliver a spoken message (P1; \texttt{session/prompt} event in ACP).
The agent processed the intent and started executing the navigation, streaming its internal reasoning and progress back to the user's interface in real time (\texttt{session/update}). 
Upon reaching the destination and before synthesizing the speech, the agent proactively paused its execution loop to request explicit human authorization (\texttt{session/request\_permission}).
Note that requiring authorization for speech synthesis was a specific implementation decision to maintain a quiet office environment, preventing the robot from emitting loud verbal messages without confirmation.
Once the user granted permission via the interface, Sancho delivered the message.
Subsequently, the user commanded the robot to return to the starting position (P2; \texttt{session/prompt}).
However, shortly after the robot started its return navigation, the user remembered an additional detail that needed to be communicated and instantly aborted the task (\texttt{session/cancel}). 
This command immediately halted the robot's physical movement.
Then, the user issued a new instruction encompassing the forgotten message (P3; \texttt{session/prompt}), prompting the agent to replan and execute the updated task.
Once completed, the user commanded the robot back to the starting location (P4; \texttt{session/prompt}). 

% Outcome
As shown, the native ACP capabilities---real-time observability, explicit authorization, and instant task interruption---allow operators to supervise and intervene at any point during the robot's operation, enforcing safety boundaries before triggering sensitive physical actions.

\begin{figure}[t]
    \centering
    \includegraphics[width=\columnwidth]{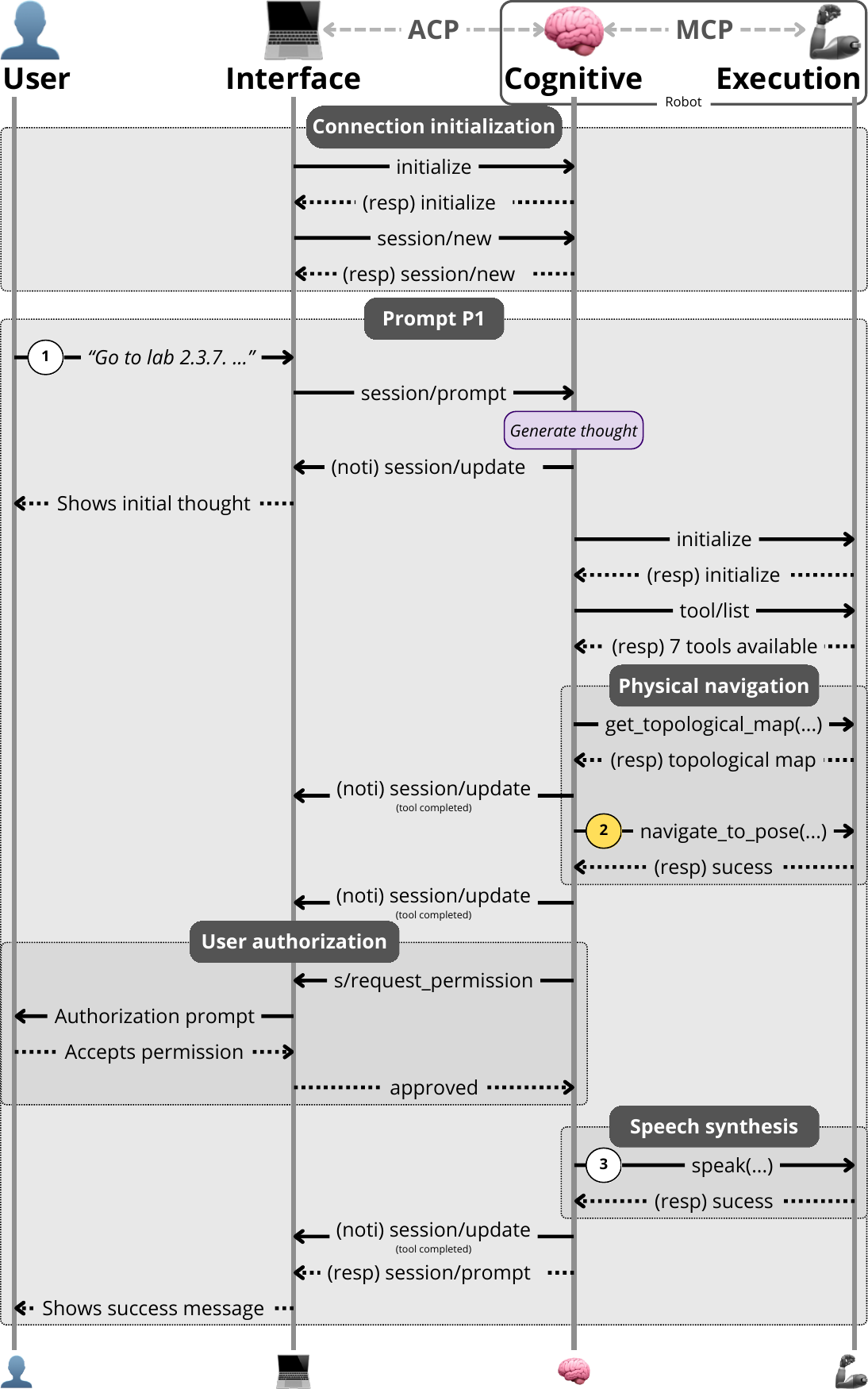}
    \caption{Sequence diagram of the user and architectural layers communication flow in the connection initialization and first interaction (P1).}
    \label{fig:p1_sequence}
    \vspace{-0.4cm}
\end{figure}

% P1 explained
To better understand the operation of the ACP and the proposed architecture, \FIG{fig:p1_sequence} illustrates the exchange across the three architectural layers during the connection initialization and the first prompt (P1).
First, the client initializes the connection by executing the \texttt{initialize} command and creating a session via a \texttt{session/\allowbreak new} request.
When the user inputs the prompt P1 (``Go to lab 2.3.7...''), the client transmits this via a \texttt{session/\allowbreak prompt} request.
The cognitive orchestrator performs an initial LLM call to generate a preliminary thought and publishes it as a progress update to the client using a \texttt{session/\allowbreak update} notification.
Next, the cognitive layer establishes its connection with the execution layer using the \texttt{initialize} and \texttt{tool/\allowbreak list} commands, discovering $7$ available MCP tools.
To perform the physical navigation, the cognitive layer first retrieves the topological map via \texttt{get\_\allowbreak topological\_\allowbreak map} and then triggers navigation via \texttt{navigate\_\allowbreak to\_\allowbreak pose}, sending progress updates to the client after each tool completes.
Upon arrival, and before executing the speech synthesis, the cognitive agent pauses its execution loop and issues an \texttt{s/\allowbreak request\_\allowbreak permission} request.
As mentioned, requiring authorization for this specific tool is an implementation decision.
Once the user accepts the permission, the client returns an \texttt{approved} response, allowing the agent to proceed by calling the \texttt{speak} tool on the execution layer.
Finally, the execution layer reports success, and the cognitive layer completes the loop by sending a final \texttt{session/\allowbreak update} notification and the \texttt{session/\allowbreak prompt} response.

\subsection{Overhead Analysis}
\label{subsec:overhead_analysis}

% Objective
To quantify the computational impact of the proposed architectural decoupling,
% What we did
we analyzed the latency breakdown of the representative usage session described in \SEC{subsec:hri_workflow_validation} as an example.
\TABLE{tab:latency_breakdown} presents the detailed timing breakdown for each interaction within the session.
These timings are categorized into five distinct phases:
(i)~\emph{Initial LLM Thought}, measuring the time taken by the LLM to generate its first thought;
(ii)~\emph{ReAct LLM Inference}, corresponding to the subsequent reasoning cycles within the ReAct loop;
(iii)~\emph{Physical Tools (MCP)}, representing the execution of hardware actions via the MCP server (navigation and speech synthesis);
(iv)~\emph{User Authorization (ACP)}, which accounts for the user response latency when granting permission; and
(v)~\emph{Protocols \& APIs}, capturing the JSON-RPC serialization, TCP round-trips, and minor software-only MCP tool calls (\eg retrieving the topological map, which do not involve physical actuation).

\begin{table}[t]
\caption{Latency breakdown of a representative experimental session}
\label{tab:latency_breakdown}
\begin{center}
\footnotesize
\setlength{\tabcolsep}{2.5pt}
\begin{tabular}{|l|c|c|c|c|c|}
\hline
\textbf{Phase} & \textbf{P1} & \textbf{P2*} & \textbf{P3} & \textbf{P4} & \textbf{Avg.} \\
\hline
Initial LLM Thought & $2.49$~s & $2.76$~s & $2.18$~s & $2.86$~s & $2.51$~s \\
ReAct LLM Inference & $5.38$~s & $1.56$~s & $3.62$~s & $4.55$~s & $4.52$~s \\
Physical Nav. (MCP) & $25.12$~s & $12.07$~s & $20.57$~s & $20.47$~s & $22.05$~s \\
Speech Synth. (MCP) & $3.94$~s & -- & $6.39$~s & $3.95$~s & $4.76$~s \\
User Auth. (ACP) & $3.02$~s & -- & $2.36$~s & $3.42$~s & $2.93$~s \\
Protocols \& APIs & $380.9$~ms & $1.1$~ms & $66.0$~ms & $68.0$~ms & $171.6$~ms \\
\hline
\textbf{Total} & \textbf{$40.33$~s} & \textbf{$16.39$~s} & \textbf{$35.19$~s} & \textbf{$35.31$~s} & \textbf{$36.94$~s} \\
\hline
\multicolumn{6}{l}{*P2 was cancelled by the user and did not complete.} \\
\multicolumn{6}{l}{Avg. represents the average of completed prompts (P1, P3, P4).}
\end{tabular}
\end{center}
\vspace{-0.4cm}
\end{table}

As shown, the completed interactions of the session required an average of $36.94$~s.
As expected, the vast majority of this time was consumed by deterministic physical navigation via ROS 2 ($22.05$~s, $59.7$\%) and LLM inference (totaling $7.03$~s, $19.0$\% across both the initial thought and the subsequent reasoning loops). 
In contrast, the communication overhead introduced by the proposed protocols is negligible. 
The ACP layer (handling session management and state streaming) introduces just $1.0$~ms ($<0.01$\%), while the MCP layer (handling connection and tool routing) adds only $170.6$~ms ($0.5$\%).
% Outcome
These results confirm that the latency footprint of dividing the system into three distinct layers, bridged by the ACP and MCP contracts, represents less than $0.5$\% of the total execution time.
Therefore, the architecture provides layer decoupling, successfully enables advanced HRI capabilities, and does not compromise the robot's real-time responsiveness.

\section{Conclusions and Limitations}
This paper has addressed a key HRI challenge in GenAI-driven robotics: the fragmentation and tight coupling between LLM-based agents and their user interfaces.
To address this limitation, we proposed the adoption of the ACP as a unifying communication protocol for the upper interface--agent link.
Combined with MCP at the lower agent--execution link, this yields a fully decoupled three-layer architecture.
The experimental evaluation demonstrated that this topology enables multiple heterogeneous user interfaces to interact with the same robotic system without requiring custom integration, while naturally supporting human-in-the-loop capabilities such as real-time intent observability, explicit authorization, and task interruption, all with negligible computational overhead.

Despite these advantages, the proposed approach presents certain limitations.
Since ACP was originally designed for coding agents in software engineering, some of its operations are not directly applicable to robotic systems.
For example, terminal process management offers limited practical utility in this context and should typically be restricted.
Furthermore, the protocol does not natively support capabilities that could be useful for robotics, such as continuous multimodal data streaming (\eg live camera feeds) or asynchronous background telemetry (\eg battery-level monitoring without explicit user prompts), requiring complementary mechanisms outside the ACP specification.

Future work may explore extensions to the ACP specification to incorporate multimodal streaming and continuous telemetry primitives, further strengthening its applicability as a unified layer for GenAI-enabled robotic systems.

\section*{Acknowledgment}
This work has been supported by the projects: MINDMAPS (PID2023-148191NB-I00), funded by the Ministry of Science, Innovation, and Universities of Spain; Voxeland (PPRO-B1-2023-017, JA.B1-09), funded by the University of Málaga; the Andalusian Plan for Research, Development, and Innovation (PPRO-TEP960-G-2023); and the Spanish FPU fellowship program under grant FPU24/02974.

\bibliographystyle{IEEEtran}
\bibliography{references}

\begin{thebibliography}{10}
\providecommand{\url}[1]{#1}
\csname url@samestyle\endcsname
\providecommand{\newblock}{\relax}
\providecommand{\bibinfo}[2]{#2}
\providecommand{\BIBentrySTDinterwordspacing}{\spaceskip=0pt\relax}
\providecommand{\BIBentryALTinterwordstretchfactor}{4}
\providecommand{\BIBentryALTinterwordspacing}{\spaceskip=\fontdimen2\font plus
\BIBentryALTinterwordstretchfactor\fontdimen3\font minus \fontdimen4\font\relax}
\providecommand{\BIBforeignlanguage}[2]{{%
\expandafter\ifx\csname l@#1\endcsname\relax
\typeout{** WARNING: IEEEtran.bst: No hyphenation pattern has been}%
\typeout{** loaded for the language `#1'. Using the pattern for}%
\typeout{** the default language instead.}%
\else
\language=\csname l@#1\endcsname
\fi
#2}}
\providecommand{\BIBdecl}{\relax}
\BIBdecl

\bibitem{siciliano2008springer}
B.~Siciliano and O.~Khatib, Eds., \emph{Springer Handbook of Robotics}.\hskip 1em plus 0.5em minus 0.4em\relax Berlin, Heidelberg: Springer, 2008.

\bibitem{firoozi2025foundation}
R.~Firoozi, J.~Tucker, S.~Tian, A.~Majumdar, J.~Sun, W.~Liu \emph{et~al.}, ``Foundation models in robotics: Applications, challenges, and the future,'' \emph{The Int. J. Robot. Res.}, vol.~44, no.~5, pp. 701--739, 2025.

\bibitem{ahn2022can}
M.~Ahn, A.~Brohan, N.~Brown, Y.~Chebotar, O.~Cortes, B.~David \emph{et~al.}, ``Do as {I} can, not as {I} say: Grounding language in robotic affordances,'' in \emph{Proc. 6th CoRL}, ser. PMLR, vol. 205.\hskip 1em plus 0.5em minus 0.4em\relax PMLR, 2023, pp. 287--318.

\bibitem{zitkovich2023rt}
B.~Zitkovich, T.~Yu, S.~Xu, P.~Xu, T.~Xiao, F.~Xia \emph{et~al.}, ``{RT-2}: Vision-language-action models transfer web knowledge to robotic control,'' in \emph{Proc. 7th CoRL}, ser. PMLR, vol. 229.\hskip 1em plus 0.5em minus 0.4em\relax PMLR, 2023, pp. 2165--2183.

\bibitem{ahn2024autort}
M.~Ahn, D.~Dwibedi, C.~Finn, M.~G. Arenas, K.~Gopalakrishnan, K.~Hausman \emph{et~al.}, ``{AutoRT}: Embodied foundation models for large scale orchestration of robotic agents,'' \emph{arXiv preprint arXiv:2401.12963}, 2024.

\bibitem{song2023llm}
C.~H. Song, J.~Wu, C.~Washington, B.~M. Sadler, W.-L. Chao, and Y.~Su, ``{LLM}-planner: Few-shot grounded planning for embodied agents with large language models,'' in \emph{Proc. ICCV}, 2023, pp. 2998--3009.

\bibitem{yao2022react}
S.~Yao, J.~Zhao, D.~Yu, N.~Du, I.~Shafran, K.~R. Narasimhan \emph{et~al.}, ``{ReAct}: Synergizing reasoning and acting in language models,'' in \emph{Proc. 11th ICLR}, 2023.

\bibitem{anthropic2024mcp}
{Anthropic}, ``{Model Context Protocol (MCP)},'' \url{https://modelcontextprotocol.io}, 2024.

\bibitem{vemprala2023chatgpt}
S.~H. Vemprala, R.~Bonatti, A.~Bucker, and A.~Kapoor, ``{ChatGPT} for robotics: Design principles and model abilities,'' \emph{IEEE Access}, vol.~12, pp. 55\,682--55\,696, 2024.

\bibitem{acp2026standard}
{Zed Industries and JetBrains}, ``{Agent Client Protocol (ACP)},'' \url{https://agentclientprotocol.com}, 2025.

\bibitem{huang2022language}
W.~Huang, P.~Abbeel, D.~Pathak, and I.~Mordatch, ``Language models as zero-shot planners: Extracting actionable knowledge for embodied agents,'' in \emph{Proc. 39th ICML}, ser. PMLR, vol. 162.\hskip 1em plus 0.5em minus 0.4em\relax PMLR, 2022, pp. 9118--9147.

\bibitem{brohan2022rt}
A.~Brohan, N.~Brown, J.~Carbajal, Y.~Chebotar, J.~Dabis, C.~Finn \emph{et~al.}, ``{RT-1}: Robotics transformer for real-world control at scale,'' in \emph{Proc. RSS}, 2023.

\bibitem{moncada2025agentic}
J.~Moncada-Ramirez, J.-L. Matez-Bandera, J.~Gonzalez-Jimenez, and J.-R. Ruiz-Sarmiento, ``Agentic workflows for improving large language model reasoning in robotic object-centered planning,'' \emph{Robotics}, vol.~14, no.~3, p.~24, 2025.

\bibitem{rana2023sayplan}
K.~Rana, J.~Haviland, S.~Garg, J.~Abou-Chakra, I.~Reid, and N.~Suenderhauf, ``{SayPlan}: Grounding large language models using {3D} scene graphs for scalable robot task planning,'' in \emph{Proc. 7th CoRL}, ser. PMLR, vol. 229.\hskip 1em plus 0.5em minus 0.4em\relax PMLR, 2023, pp. 23--72.

\bibitem{huang2022inner}
W.~Huang, F.~Xia, T.~Xiao, H.~Chan, J.~Liang, P.~Florence \emph{et~al.}, ``Inner monologue: Embodied reasoning through planning with language models,'' in \emph{Proc. 6th CoRL}, ser. PMLR, vol. 205.\hskip 1em plus 0.5em minus 0.4em\relax PMLR, 2023, pp. 1769--1782.

\bibitem{liang2023code}
J.~Liang, W.~Huang, F.~Xia, P.~Xu, K.~Hausman, B.~Ichter \emph{et~al.}, ``Code as policies: Language model programs for embodied control,'' in \emph{Proc. ICRA}.\hskip 1em plus 0.5em minus 0.4em\relax IEEE, 2023, pp. 9493--9500.

\bibitem{singh2022progprompt}
I.~Singh, V.~Blukis, A.~Mousavian, A.~Goyal, D.~Xu, J.~Tremblay \emph{et~al.}, ``{ProgPrompt}: Generating situated robot task plans using large language models,'' in \emph{Proc. ICRA}.\hskip 1em plus 0.5em minus 0.4em\relax IEEE, 2023.

\bibitem{guan2025roboneuron}
W.~Guan, Q.~Hu, H.~Xi, C.~Zhang, A.~Li, and J.~Cheng, ``{RoboNeuron}: A middle-layer infrastructure for agent-driven orchestration in embodied {AI},'' \emph{arXiv preprint arXiv:2512.10394}, 2025.

\bibitem{ros2mcp_community}
{robotmcp Contributors}, ``{ROS MCP Server},'' \url{https://github.com/robotmcp/ros-mcp-server}, 2025.

\bibitem{ren2023robots}
A.~Z. Ren, A.~Dixit, A.~Bodrova, S.~Singh, S.~Tu, N.~Brown \emph{et~al.}, ``Robots that ask for help: Uncertainty alignment for large language model planners,'' in \emph{Proc. 7th CoRL}, ser. PMLR, vol. 229.\hskip 1em plus 0.5em minus 0.4em\relax PMLR, 2023.

\bibitem{cui2023no}
Y.~Cui, S.~Karamcheti, R.~Palleti, N.~Shivakumar, P.~Liang, and D.~Sadigh, ``No, to the right: Online language corrections for robotic manipulation via shared autonomy,'' in \emph{Proc. HRI}, 2023, pp. 93--101.

\end{thebibliography}

\end{document}